\documentclass{article}
\pdfoutput=1
\usepackage[final]{neurips_2021}
\usepackage{amsmath, amsthm, amsfonts}
\usepackage{graphicx}
\usepackage[hidelinks]{hyperref}
\usepackage{wrapfig}
\graphicspath{{figures/}}

\newcommand{\hmu}{\hat{\mu}}

\newcommand{\tests}{\varphi}
\newcommand{\ttests}{\Phi}

\newcommand{\E}{\mathbf{E}}
\newcommand{\ind}{\mathbf{1}}
\newcommand{\eps}{\epsilon}

\DeclareMathOperator{\argmax}{\textnormal{argmax}}

\newcommand{\betad}{\text{beta}}
\newcommand{\Bin}{\textnormal{Bin}}

\author{
    Ben Chugg \\
    Stanford University \\ 
    \texttt{benchugg@law.stanford.edu}
    \And 
    Daniel E. Ho \\
    Stanford University \\ 
    \texttt{deho@law.stanford.edu}
}

\title{Reconciling Risk Allocation and Prevalence Estimation in Public Health Using Batched Bandits}
\begin{document}

\maketitle

\begin{abstract}
    In many public health settings, there is a perceived tension between allocating resources to known vulnerable areas and learning about the overall prevalence of the problem. Inspired by a door-to-door Covid-19 testing program we helped design, 
    we combine multi-armed bandit strategies and insights from sampling theory to demonstrate how to recover accurate prevalence estimates while continuing to allocate resources to at-risk areas. We use the outbreak of an infectious disease as our running example.  
    The public health setting has several characteristics distinguishing it from typical bandit settings, such as distribution shift (the true disease prevalence is changing with time) and batched sampling (multiple decisions must be made simultaneously). Nevertheless, we demonstrate that several bandit algorithms are capable out-performing greedy resource allocation strategies, which often perform worse than random allocation as they fail to notice outbreaks in new areas. 
\end{abstract}


 
\section{Introduction}

Public health crises --- such as lead paint, drinking water contamination, natural disasters, and infectious disease outbreaks --- demand effectively distributing scarce resources. In 
 the early stages of the COVID-19 pandemic, for instance, public health officials grappled with how to best allocate diagnostic tests~\cite{mina2021covid}.
Some advocate random sampling to learn about the  disease~\cite{padula2020only,muller2020testing}, while others favor more targeted allocation~\cite{cleevely2020workable,chugg2021evaluation}. This is perceived to pose a tradeoff: Random sampling is seen to enable reliable prevalence estimates and the discovery of new hot spots, while risk targeting focuses on symptomatic and/or exposed individuals to ostensibly drive down disease rates.\footnote{There are various mechanisms accounting for why uncovering infected individuals reduces disease rates, e.g., changing behavior and enabling contract tracing~\cite{kasy2020adaptive,piguillem2020optimal}.} While the general tension runs throughput public health, the setting presents novel challenges to classical bandit approaches. Outbreaks present \emph{distribution shifts} and allocation is necessarily \emph{batched} for operational reasons. 

In this short work, we demonstrate that these two objectives are not in zero-sum conflict. Inspired by a door-to-door Covid-19 testing program we helped design, we demonstrate that the batched setting allows us to  recover unbiased prevalence estimates and we investigate the performance of bandits in the presence of distribution shift.  To illustrate, we 
consider the early stages of an infectious disease outbreak when tests are not widely available. A public health agency must decide how to allocate tests between $K$ different units, $U_1,\dots,U_K$. Units might be neighbourhoods~\cite{chugg2021evaluation}, apartments buildings~\cite{lai2004understanding}, or dorm rooms on a college campus~\cite{wilson2020multiple}. They might also be artificially constructed clusters of individuals based on interaction networks~\cite{prasse2021clustering}.  
We consider a series of discrete timesteps $t=1,\dots,T$, which could represent days or weeks. 
At time $t$, $m_t$ tests can be distributed among the regions in any way we choose. The supply may vary from period to period, in which case $m_t\neq m_{t+1}$.

Our approach involves treating each unit as an arm in the multi-armed bandit problem~\cite{slivkins2019introduction}. The prevalence rate\footnote{true proportion of infected individuals} in each unit is the ``reward'' distribution, thus encouraging tests to be allocated to units with higher proportions of infected individuals. Unlike most bandit problems our reward distributions are \emph{non-stationary}~\cite{besbes2014stochastic}. In particular, they change as a function of the disease spread and the number of tests (Section~\ref{sec:model}). In addition, we must make take multiple actions at once (actions are \emph{batched})~\cite{huang2016linear}. We investigate four bandit algorithms: Upper confidence bound sampling, Thompson sampling, Exponential Weights (Exp3), and $\epsilon$-greedy. We discuss their details and how they've been extended to the non-stationary and batched setting in Section~\ref{sec:strategies}. The following section discusses how we model the spread of the disease in each unit and the effect of testing. Code is available at \url{https://github.com/reglab/mab-infectious-disease}. 


\section{Disease Model}
\label{sec:model}

Many  mathematical models for  infectious disease dynamics exist~\cite{chowell2016mathematical}.  Here, we use a simple logistic model which incorporates the effect of testing and contact tracing. We note, however, that none of the bandit sampling strategies nor the prevalence estimation rely on the specifics of the disease model. 

Given a unit with population $N$, let $C(t)$ be the number of disease cases at time $t$. 
We begin by considering the logistic growth equation, $f'(t) = \alpha f(t)(1 - f(t)/N)$, which accounts for the carrying capacity, $N$, of the population and the \emph{growth rate} $\alpha$ of the virus. In our setting, however, we want to take into account the additional effects of testing. For an empirically chosen parameter $\beta$, we thus add an extra term to the logistic equation and obtain a Bernoulli type differential equation: 

\begin{equation}
\label{eq:growth}
  \frac{dC(t)}{dt} = \alpha C(t) \bigg( 1 - \beta \tests(t)-\frac{C(t)}{N}\bigg),  
\end{equation}

where $\tests(t)$ is the number of tests conducted at time $t$.  The additional term $\beta\tests(t)$ represents the impact on the rate of growth of the virus: The more tests performed at time $t$, the lower the rate of growth. We assume that the term $\beta\tests(t)$ captures not only the immediate effects of testing, but the longer run effects of subsequent interventions, such as isolation / quarantine and contact tracing. See Figure~\ref{fig:tests_and_improvement} for an illustration of the model dynamics and the effects of testing. In our setting, disease spread in each unit $U_i$ evolves independently according to Equation~\eqref{eq:growth}.

\section{Bandit Strategies}
\label{sec:strategies}

At each time $t$, a given strategy chooses $m_t$ regions to sample, and then samples individuals uniformly \emph{within} that region. In order to adapt the strategies to the non-stationary setting, we apply a discount factor $\gamma\in(0,1)$ to the information received over time~\cite{cavenaghi2021non,russac2019weighted}. Let $\alpha_\tau$ be the number of positive tests witnessed at time $\tau$. If making a decision based on the overall number of, say, positive tests seen from time 0 to $t$, each strategy will use the discounted quantity $\sum_{\tau=0}^t\gamma^{t-\tau} a_t$. In this way, each strategy smoothly forgets the past.  

We give a very brief overview of each strategy. For more details, we refer readers to~\cite{slivkins2019introduction}. 

\paragraph{$\eps$-greedy.}
Perhaps the simplest bandit strategy is $\eps$-greedy, which chooses a random action with probability $\eps$ and chooses the action with the highest expected reward (i.e., the positivity rate, meaning the fraction of test results yielding a positive test result) otherwise. In the batched setting, we simply repeat this process $m_t$ times. 

\paragraph{Thompson Sampling.}
Thompson sampling maintains a beta distribution over each unit, $\betad(P_k,N_k)$ where $P_k$ (resp., $N_k$) is the (discounted) number positive (resp., negative) tests in unit $k$. To batch sample, first draw $p_k^1,\dots,p_k^{m_t}\sim\betad(P_k,N_k)$ for each $k=1,\dots,K$, and then for each $r=1,\dots,m_t$ sample region $\argmax_{k}p_k^r$.

\paragraph{UCB Sampling.}
We use a binomial proportion confidence interval as our confidence bound. Formally, we use a Clopper-Pearson interval, so unit $k$ receives a score 
\begin{equation*}
  U(k,t) = \sup_\theta \{\theta|\Pr(\Bin(P_k+N_k,\theta)\leq P)\geq \alpha/2\}.  
\end{equation*}
In order to batch sample, we form a distribution over the regions using their scores: 
\begin{equation}
\label{eq:prob_ucb}
  \pi_k=\frac{U(k,t)}{\sum_{k'}U(k',t)},  
\end{equation}
and sample from this distribution (with replacement) $m_t$ times.

\paragraph{Exp3.}
Each unit $i$ begins with weight $w_i(0)=1$. At time $t$, we sample unit $i$ with probability 
\begin{equation}
\label{eq:prob_exp3}
  \pi_k(t) = (1-\eps)\frac{w_k(t)}{\sum_{k'}w_{k'}(t)} + \frac{\eps}{K},  
\end{equation}
for some $\eps\in(0,1)$.
If unit $k$ was sampled and $r_k$ is its reward, we update as the weight as  
\[w_k(t+1)=w_k(t)\exp\bigg(\frac{\eps r_k}{\pi_k(t)K}\bigg).\]
Otherwise, we set $w_k(t+1)=w_k(t)$. For us, the reward $r_k$ is the number of positive tests received after testing unit $k$.

\section{Estimating Prevalence}
\label{sec:population_estimation}
For an individual $i$ at time $t$, let $y_t(i)\in\{0,1\}$ indicate whether they are infected. Write $i\in U_k$ if individual $i$ is in unit $U_k$. 
We're searching for an estimate of the true prevalence---a \emph{population estimate}---of the disease at time $t$. That is, 
\begin{equation}
\label{eq:total_pos}
    \mu_t = \frac{1}{\sum_k N_k}\sum_{k=1}^K\sum_{i\in U_k} y_t(i). 
\end{equation}
\begin{figure}[t]
    \centering
    \begin{minipage}{0.48\textwidth}
    \includegraphics[scale=0.3]{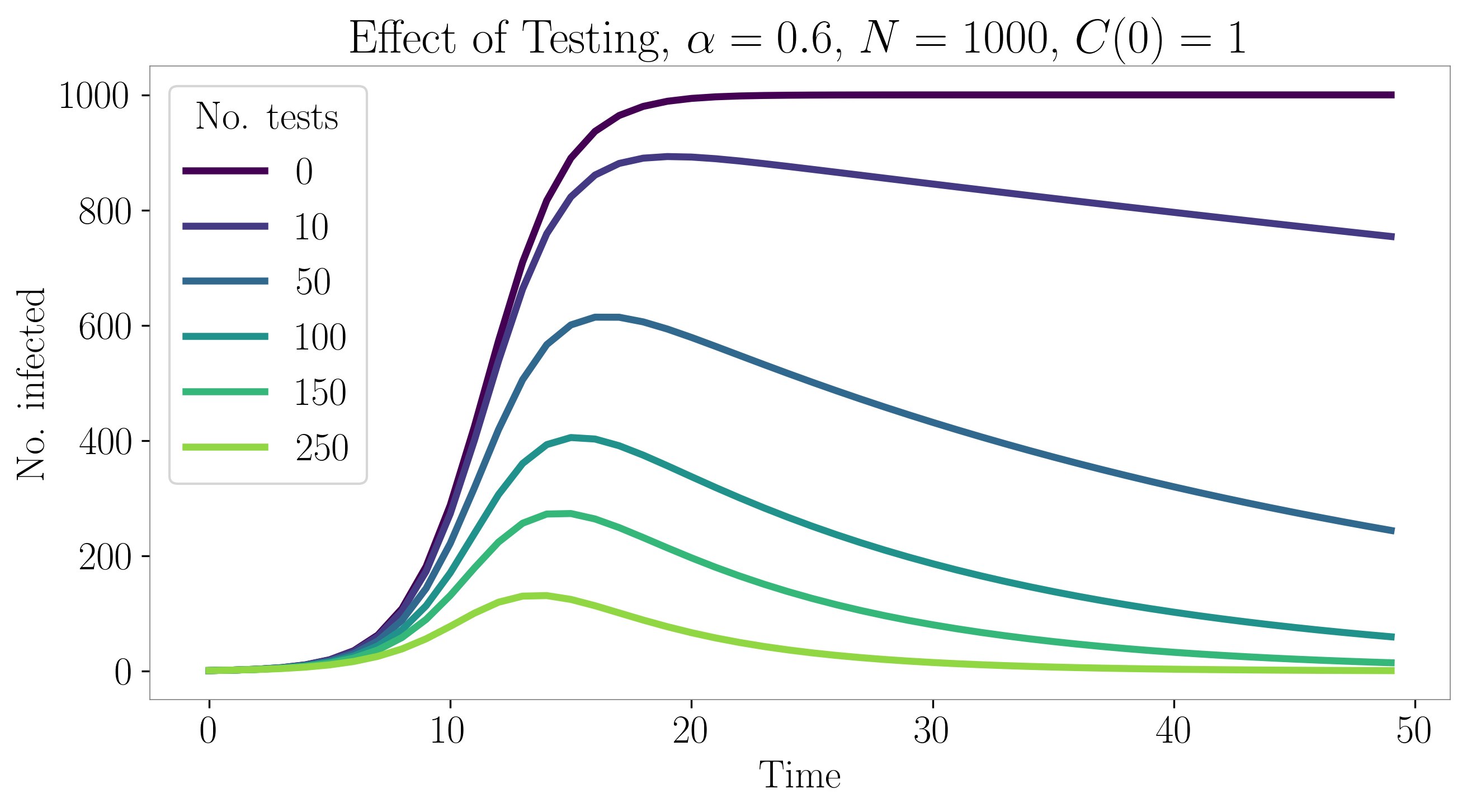}
    \end{minipage}
    \hfill
    \begin{minipage}{0.48\textwidth}
    \includegraphics[scale=0.3]{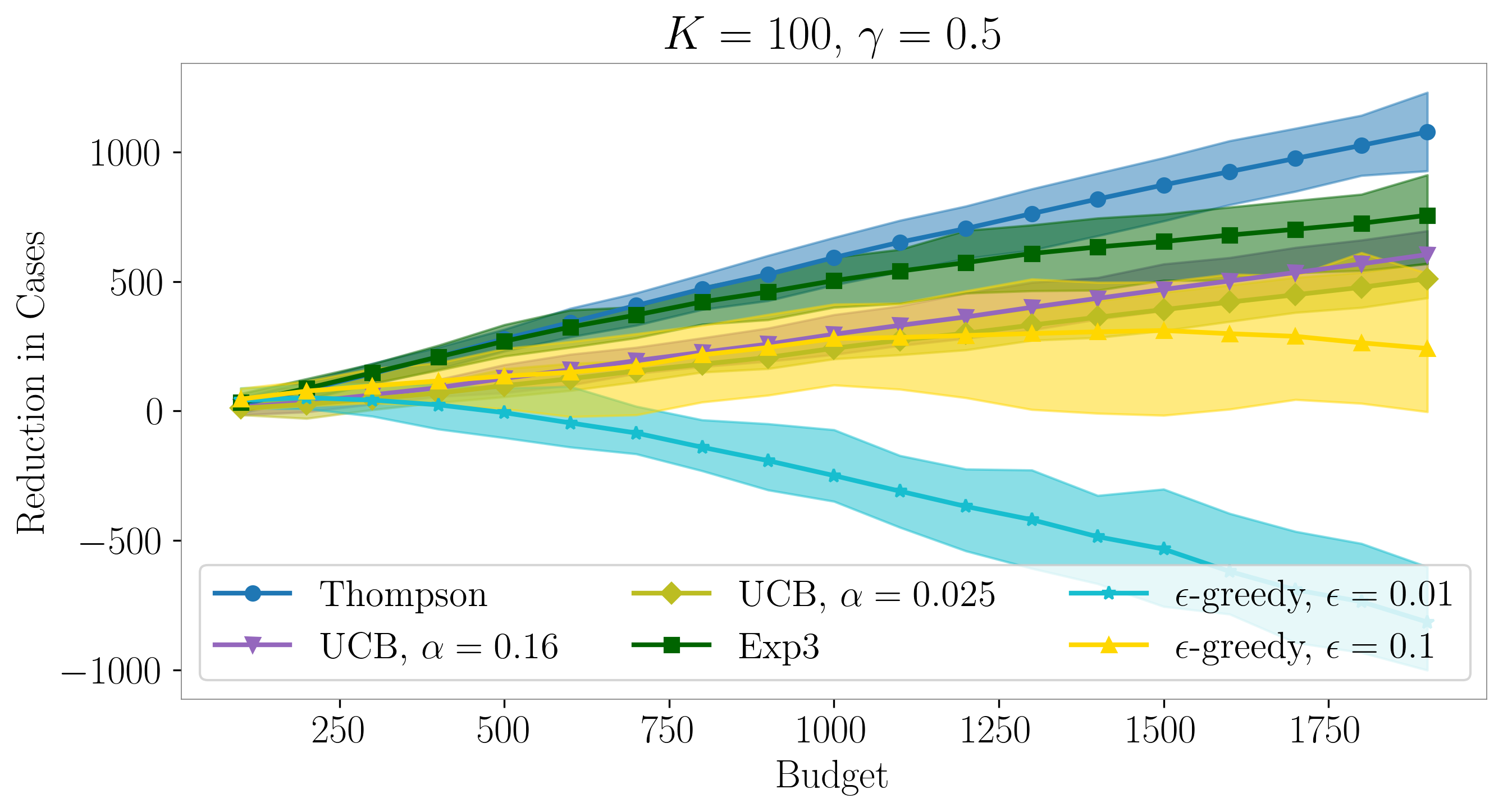}
    \end{minipage}
    \caption{\textbf{Left:} Effect of testing on the spread of the disease. The legend indicates how many tests were administered at each timestep. \textbf{Right:} The reduction in number of positive cases, plotted with 68\% confidence intervals. For each budget, the simulation was run for 30 timesteps and repeated 30 times. Each region had a population of $N=1000$, and $c(0)$ and $\alpha$ were randomly selected uniformly at random from $\{1,\dots,20\}$ and $(0,1)$ respectively, independently for each region. All experiments use $\beta=0.001$.}
    \label{fig:tests_and_improvement}
\end{figure}

We employ a Horvitz-Thompson-type approach which takes into account the probability that each region was sampled~\cite{horvitz1952generalization}.  
Put $N=\sum_k N_k$ and let $S_t$ be the set of individuals tested during timestep $t$. Let $\pi_k(t)$ denote the probability that unit $U_k$ was sampled by the bandit strategy (e.g., Equations~\eqref{eq:prob_ucb} or \eqref{eq:prob_exp3}).\footnote{Note that $\pi_k(t)$ is the probability that unit $k$ is selected once---to form $S_t$ we sample from the distribution $\{\pi_k(t)\}$ a total of $m_t$ times.} The quantity
\begin{equation}
    \label{eq:mu_hat}
    \hmu_t \equiv \frac{1}{Nm_t}\sum_{k}\frac{N_k}{\pi_k(t)}\sum_{i\in U_k\cap S_t}y_t(i),
\end{equation}
is an unbiased estimate of $\mu_t$ (see Appendix~\ref{app:unbiased} for a proof).\footnote{In this work, we assume for simplicity that each the false positive and false negative rates are zero, so that when individual $i$ is tested, we accurately receive $y_t(i)$. In reality, we receive $\hat{y}_t(i)$, which is sometimes equal to $y_t(i)$ but can be different for a false positive or false negative test. If the false positive and negative rates were known however, the estimate could be corrected to take this into account.} 


\section{Experiments}
\label{sec:experiments}

The right hand side of Figure~\ref{fig:tests_and_improvement} examines the improvement of each bandit strategy over random selection, using a discount rate of $\gamma=0.5$. For each budget, we run the simulation for 30 timesteps. Each point reflects the difference between the total number of positive cases across all regions \emph{at the end of the 30 timesteps}, between the bandit strategy and random sampling. 
Each bandit strategy causes a decrease in the total number of positive cases when compared with random selection, with the exception of $\epsilon$-greedy at $\eps=0.01$ (i.e., using 1\% of its budget to perform random exploration). 
Thompson Sampling and Exp3 are the two best performing strategies, especially as the budget grows. 

The results of more experiments across a wider range of parameters are shown in Appendix~\ref{app:experiments}. The general trends are that (i) as the number of allocation units $K$ decreases, the differences between different strategies become less apparent, and (ii) as the budget increases and/or $K$ increases, Thompson Sampling, Exp3, and UCB outperform $\eps$-greedy by wider margins.

\begin{figure}[t]
    \centering
    \includegraphics[scale=0.4]{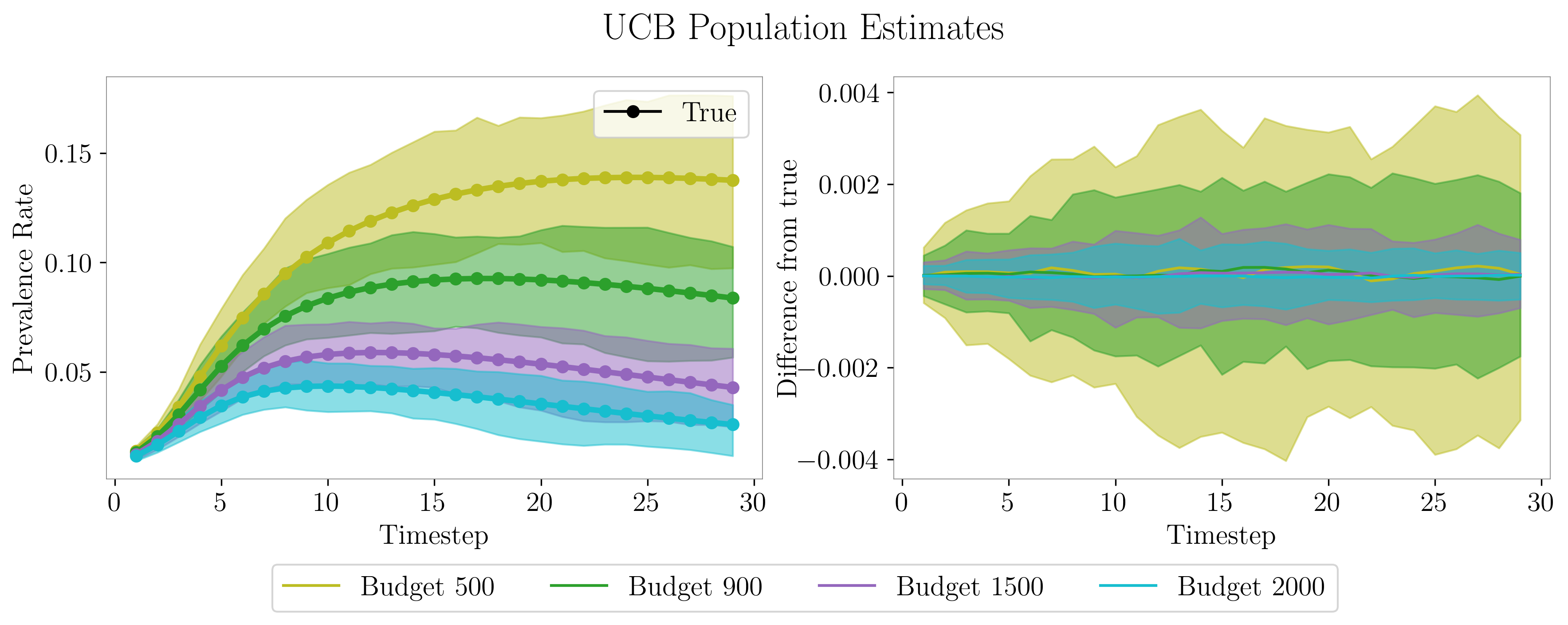}
    \caption{Population estimation as a function of the budget. \textbf{Left:} The dotted lines are the average true mean across 50 runs, while the  fill captures the set of all estimates. \textbf{Right:} The difference between the true mean and the estimated mean. Both graphs use $K=10$, 30 timesteps, and $\gamma=0.5$.}
    \label{fig:pop_est}
\end{figure}

Using UCB as an example, Figure~\ref{fig:pop_est} illustrates population estimation as a function of the budget. The figure on the left hand side demonstrates the evolution of the true prevalence rates (across all units) and the estimates over time. Note that the true means are different for different budgets because testing changes the prevalence of the disease, and a higher budget implies more testing. Thus, a lower budget results in a higher prevalence. The right hand side of Figure~\ref{fig:pop_est} demonstrates the effect of different budget sizes on the variance of the estimate. While the estimates are unbiased on average across all three budgets, smaller budgets result in larger variance.  Of course, this is due to the fact that the variance of $\hmu_t$ grows as a function of the inverse of $p_ip_j$, where $p_i$ is the probability that individual $i$ is tested. This, in turn, is a function of $m_t$. Thus, as $m_t\to0$, $Var(\hmu_t)\to\infty$. 

\section{Conclusions}
This work has taken initial steps towards understanding both how multi-armed bandits can enable resources to be more effectively distributed during a public health crisis, and demonstrated that allocating resources to high-risk areas while maintaining an accurate understanding of the true dynamics of the problem (in this case, the spread of an infectious disease) are not necessarily in conflict. 

\bibliographystyle{plain}
\bibliography{main.bib}

\begin{thebibliography}{10}

\bibitem{besbes2014stochastic}
Omar Besbes, Yonatan Gur, and Assaf Zeevi.
\newblock Stochastic multi-armed-bandit problem with non-stationary rewards.
\newblock {\em Advances in neural information processing systems}, 27:199--207,
  2014.

\bibitem{cavenaghi2021non}
Emanuele Cavenaghi, Gabriele Sottocornola, Fabio Stella, and Markus Zanker.
\newblock Non stationary multi-armed bandit: Empirical evaluation of a new
  concept drift-aware algorithm.
\newblock {\em Entropy}, 23(3):380, 2021.

\bibitem{chowell2016mathematical}
Gerardo Chowell, Lisa Sattenspiel, Shweta Bansal, and C{\'e}cile Viboud.
\newblock Mathematical models to characterize early epidemic growth: A review.
\newblock {\em Physics of life reviews}, 18:66--97, 2016.

\bibitem{chugg2021evaluation}
Ben Chugg, Lisa Lu, Derek Ouyang, Benjamin Anderson, Raymond Ha, Alexis
  D’Agostino, Anandi Sujeer, Sarah~L Rudman, Analilia Garcia, and Daniel~E
  Ho.
\newblock Evaluation of allocation schemes of covid-19 testing resources in a
  community-based door-to-door testing program.
\newblock In {\em JAMA Health Forum}, volume~2, pages e212260--e212260.
  American Medical Association, 2021.

\bibitem{cleevely2020workable}
Matthew Cleevely, Daniel Susskind, David Vines, Louis Vines, and Samuel Wills.
\newblock A workable strategy for covid-19 testing: stratified periodic testing
  rather than universal random testing.
\newblock {\em Oxford Review of Economic Policy}, 36(Supplement\_1):S14--S37,
  2020.

\bibitem{horvitz1952generalization}
Daniel~G Horvitz and Donovan~J Thompson.
\newblock A generalization of sampling without replacement from a finite
  universe.
\newblock {\em Journal of the American statistical Association},
  47(260):663--685, 1952.

\bibitem{huang2016linear}
Kuan-Hao Huang and Hsuan-Tien Lin.
\newblock Linear upper confidence bound algorithm for contextual bandit problem
  with piled rewards.
\newblock In {\em Pacific-Asia Conference on Knowledge Discovery and Data
  Mining}, pages 143--155. Springer, 2016.

\bibitem{kasy2020adaptive}
Maximilian Kasy and Alexander Teytelboym.
\newblock Adaptive targeted infectious disease testing.
\newblock {\em Oxford Review of Economic Policy}, 36(Supplement\_1):S77--S93,
  2020.

\bibitem{lai2004understanding}
PC~Lai, CM~Wong, AJ~Hedley, SV~Lo, PY~Leung, J~Kong, and GM~Leung.
\newblock Understanding the spatial clustering of severe acute respiratory
  syndrome (sars) in hong kong.
\newblock {\em Environmental Health Perspectives}, 112(15):1550--1556, 2004.

\bibitem{mina2021covid}
Michael~J Mina and Kristian~G Andersen.
\newblock Covid-19 testing: One size does not fit all.
\newblock {\em Science}, 371(6525):126--127, 2021.

\bibitem{muller2020testing}
Markus M{\"u}ller, Peter~M Derlet, Christopher Mudry, and Gabriel Aeppli.
\newblock Testing of asymptomatic individuals for fast feedback-control of
  covid-19 pandemic.
\newblock {\em Physical biology}, 17(6):065007, 2020.

\bibitem{padula2020only}
William~V Padula.
\newblock Why only test symptomatic patients? consider random screening for
  covid-19, 2020.

\bibitem{piguillem2020optimal}
Facundo Piguillem and Liyan Shi.
\newblock Optimal covid-19 quarantine and testing policies.
\newblock 2020.

\bibitem{prasse2021clustering}
Bastian Prasse, Karel Devriendt, and Piet Van~Mieghem.
\newblock Clustering for epidemics on networks: a geometric approach.
\newblock {\em Chaos: An Interdisciplinary Journal of Nonlinear Science},
  31(6):063115, 2021.

\bibitem{russac2019weighted}
Yoan Russac, Claire Vernade, and Olivier Capp{\'e}.
\newblock Weighted linear bandits for non-stationary environments.
\newblock {\em arXiv preprint arXiv:1909.09146}, 2019.

\bibitem{slivkins2019introduction}
Aleksandrs Slivkins.
\newblock Introduction to multi-armed bandits.
\newblock {\em arXiv preprint arXiv:1904.07272}, 2019.

\bibitem{wilson2020multiple}
Erica Wilson, Catherine~V Donovan, Margaret Campbell, Thevy Chai, Kenneth
  Pittman, Arlene~C Se{\~n}a, Audrey Pettifor, David~J Weber, Aditi Mallick,
  Anna Cope, et~al.
\newblock Multiple covid-19 clusters on a university campus—north carolina,
  august 2020.
\newblock {\em Morbidity and Mortality Weekly Report}, 69(39):1416, 2020.

\end{thebibliography}

\appendix 

\section{Solving Equation~\eqref{eq:growth}}
\label{app:growth}
Let $\ttests(t)=\int_0^t \tests(z)dz$ be the total number of tests conducted in region $r$ up to and including time $t$. The solution to Equation \eqref{eq:growth} is given by  

\begin{equation}
\label{eq:growth_soln}
    C(t) = 
    \frac{C(0)N\lambda_{\alpha,\beta}(t)}
    {N-C(0) + C(0)\alpha\int_0^t \lambda_{\alpha,\beta}(\tau)d\tau},
\end{equation}
where 
\begin{equation*}
  \lambda_{\alpha,\beta}(x) = \exp(\alpha(x - \ttests(x))).  
\end{equation*}

Observe that for $\tests\equiv 0$, the solution reduces to the solution of the usual logistic growth equation. For $\tests\not\equiv0$, however, the integral in the denominator of $C(t)$ is harder to evaluate owing to the term $\ttests(t)$.  
To evaluate it, we recall that $\ttests(t)$ is a function of discrete time steps: it changes only at times integer times $t=0,1,\dots$ and is constant on the interval $[t,t+1)$. Thus, 
\begin{align*}
    \int_0^t \lambda_{\alpha,\beta}(x)dx &= \sum_{\tau=0}^{\lfloor t\rfloor -1}\int_{\xi=\tau}^{\xi=\tau+1}\exp(\alpha(\xi-\ttests(\tau)))d\xi + \int_{\lfloor t\rfloor }^t \exp(\alpha(\xi-\ttests(\lfloor t\rfloor)))d\xi \\ 
    &= \frac{1}{\alpha}\sum_{\tau=0}^{\lfloor t\rfloor -1}e^{-\alpha \ttests(\tau)}(e^{\alpha(\tau+1)}-e^{\alpha\tau}) + \frac{1}{\alpha}e^{-\alpha  \ttests(\lfloor t\rfloor)}(e^{\alpha t}-e^{\alpha\lfloor t\rfloor }),
\end{align*}
which is straightforward to evaluate computationally.

\section{$\hmu_t$ is Unbiased}
\label{app:unbiased}
This is straightforward to see after noting that for an individual $i$ in unit $U_{k_i}$, $\Pr(i\in S_t)=m_t\pi_{k_i}(t)/N_{k_i}$. We can thus take the expectation of \eqref{eq:mu_hat} and rearrange it to read 
\begin{align*}
    \E[\hmu_t] &= \frac{1}{Nm_t}\E\bigg\{\sum_k N_k\pi_k(t)^{-1}\sum_{i=1}^N y_t(i) \ind(i\in U_k)\ind (i \in S_t)\bigg\} \\ 
    &=\frac{1}{Nm_t}\E\bigg\{\sum_{i=1}^N y_t(i) \sum_k N_k \pi_k(t)^{-1} \ind(i\in U_k)\ind(i\in S_t) \bigg\}\\
    &= \frac{1}{Nm_t}\E\bigg\{\sum_{i=1}^N y_t(i) N_{k_i}\pi_{k_i}(t)^{-1}\ind(i\in S_t)\bigg\} \\ 
    &=
    \frac{1}{Nm_t}\sum_{i=1}^N y_t(i) N_{k_i} \pi_{k_i}(t)^{-1} \Pr(i\in S_t) = \frac{1}{N}\sum_{i=1}^N y_t(i).
\end{align*}
Note that the result for $\Pr(i\in S_t)$ is because
$|U_{k_i}\cap S|=k$ is distributed as a binomial. So 
\begin{align*}
    \Pr(i \in S_t) 
    &=\sum_{\ell=1}^{m_t} \Pr(i\in S_t||U_{k_i}\cap S_t|=\ell)\Pr(|U_{k_i}\cap S|=\ell)\\
    &=\sum_{\ell=1}^{m_t}\frac{\ell}{N_{k_i}}{m_t\choose \ell}\pi_{k_i}^\ell (1-\pi_{k_i})^{m_t-\ell} \\ 
    &= \frac{m_t\pi_{k_i}}{N_{k_i}}.
\end{align*}

\newpage
\section{Additional Experiments}
\label{app:experiments}

\begin{figure}[h]
\begin{minipage}{1\textwidth}
\includegraphics[scale=0.5]{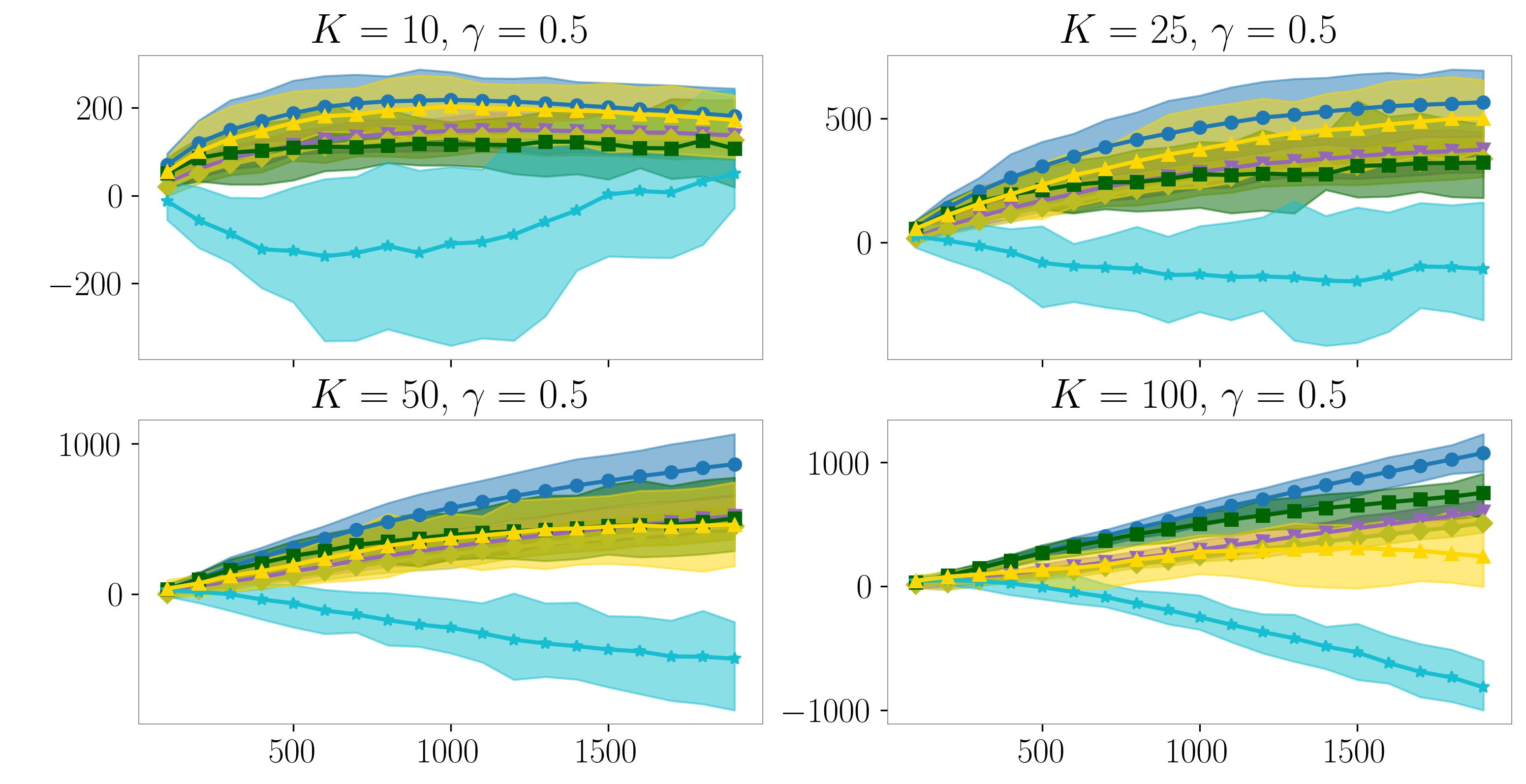}
\end{minipage}
\begin{minipage}{1\textwidth}
\includegraphics[scale=0.5]{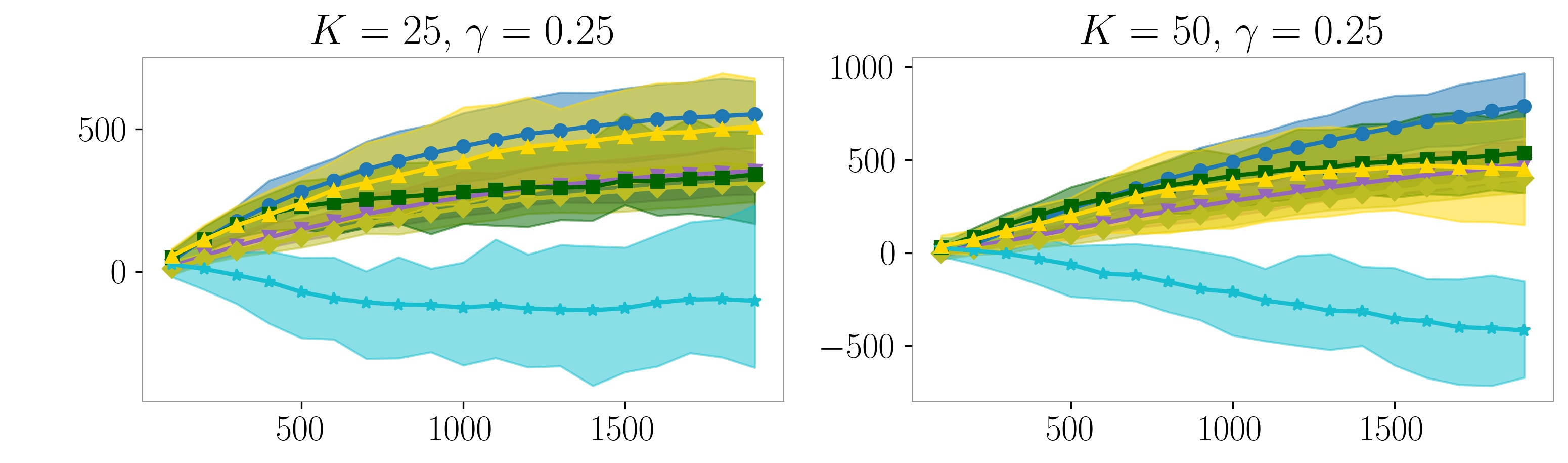}
\end{minipage}
\begin{minipage}{1\textwidth}
\includegraphics[scale=0.5]{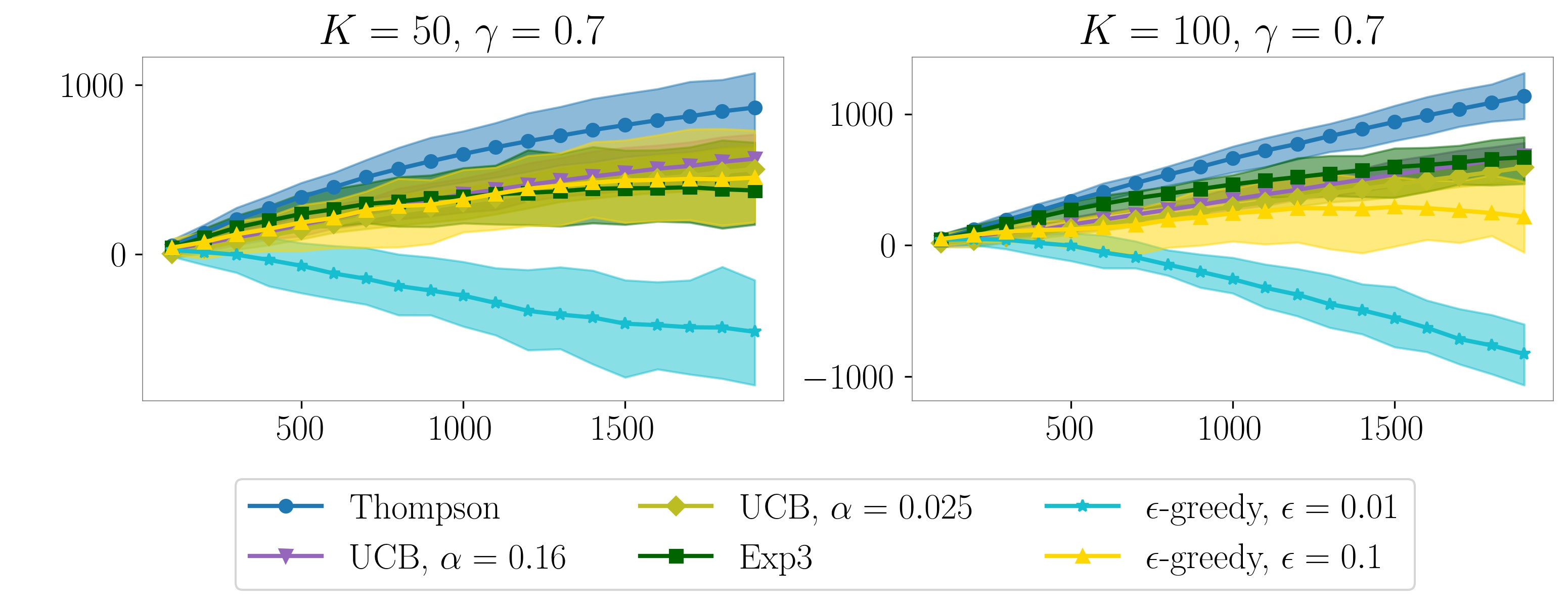}
\end{minipage}
    \caption{Improvement over random sampling. As in the right hand side of Figure~\ref{fig:tests_and_improvement}, the $x$-axis is the budget, and the $y$-axis is the difference in number of cases between the given strategy and random sampling by the end of 30 timesteps.  As $K$ increases, we see a general increase in the effectiveness of Thompson Sampling and Exp3 relative to the two $\eps$-greedy strategies. For all values of $K$ and $\gamma$, $\eps$-greedy at $\eps=0.01$ performs worse than random sampling across the majority of budgets. This suggests that with such a small exploration budget, it is unable to discover those units whose prevalence is increasing drastically over time. Interestingly, at lower values of $K$, $\eps$-greedy with more exploration (10\%) is unable to compete with the other bandit algorithms.  }
    \label{fig:extra_experiments}
\end{figure}

\end{document}